\documentclass[letterpaper]{article} 
\usepackage{aaai2026}  
\usepackage{times}  
\usepackage{helvet}  
\usepackage{courier}  
\usepackage[hyphens]{url}  
\usepackage{graphicx} 
\usepackage{natbib}  
\usepackage{caption} 
\usepackage{algorithm}
\usepackage{algorithmic}

\usepackage{enumitem}
\usepackage{amsmath}
\usepackage{booktabs} 
\usepackage{multirow}
\usepackage{tabularx}
\usepackage{pifont}   
\usepackage{amsthm}
\usepackage{booktabs} 
\usepackage{multirow} 
\usepackage{lipsum} 
\usepackage{booktabs}
\usepackage{tabularx}
\usepackage{adjustbox}
\usepackage{booktabs} 
\usepackage{newfloat}
\usepackage{listings}
\usepackage{newfloat}
\usepackage{listings}
\DeclareCaptionStyle{ruled}{labelfont=normalfont,labelsep=colon,strut=off} 
\floatstyle{ruled}
\newfloat{listing}{tb}{lst}{}
\floatname{listing}{Listing}
\title{PIMRL: Physics-Informed Multi-Scale Recurrent Learning for Burst-Sampled Spatiotemporal Dynamics}

\author{
  Han Wan\equalcontrib,
  Qi Wang\equalcontrib,
  Yuan Mi,
  Rui Zhang\thanks{Corresponding authors: Rui Zhang and Hao Sun.},
  Hao Sun$^\dag$
}
\affiliations{
  Gaoling School of Artificial Intelligence, Renmin University of China, Beijing, China\\
  \{wanhan2001, 2021000184, miyuan, rayzhang, haosun\}@ruc.edu.cn
}

\begin{document}

\maketitle

\begin{abstract}

Deep learning has shown strong potential in modeling complex spatiotemporal dynamics. However, most existing methods depend on densely and uniformly sampled data, which is often unavailable in practice due to sensor and cost limitations. In many real-world settings, such as mobile sensing and physical experiments, data are burst-sampled with short high-frequency segments followed by long gaps, making it difficult to learn accurate dynamics from sparse observations. To address this issue, we propose Physics-Informed Multi-Scale Recurrent Learning (PIMRL), a novel framework specifically designed for burst-sampled spatiotemporal data. PIMRL combines macro-scale latent dynamics inference with micro-scale adaptive refinement guided by incomplete prior information from partial differential equations (PDEs). It further introduces a temporal message-passing mechanism to effectively propagate information across burst intervals. This multi-scale architecture enables PIMRL to model complex systems accurately even under severe data scarcity. We evaluate our approach on five benchmark datasets involving 1D to 3D multi-scale PDEs. The results show that PIMRL consistently outperforms state-of-the-art baselines, achieving substantial improvements and reducing errors by up to 80\% in the most challenging settings, which demonstrates the clear advantage of our model. Our work demonstrates the effectiveness of physics-informed recurrent learning for accurate and efficient modeling of sparse spatiotemporal systems.

\end{abstract}

\section{Introduction}

\begin{figure} 
  \centering
  \includegraphics[width=0.4\textwidth]{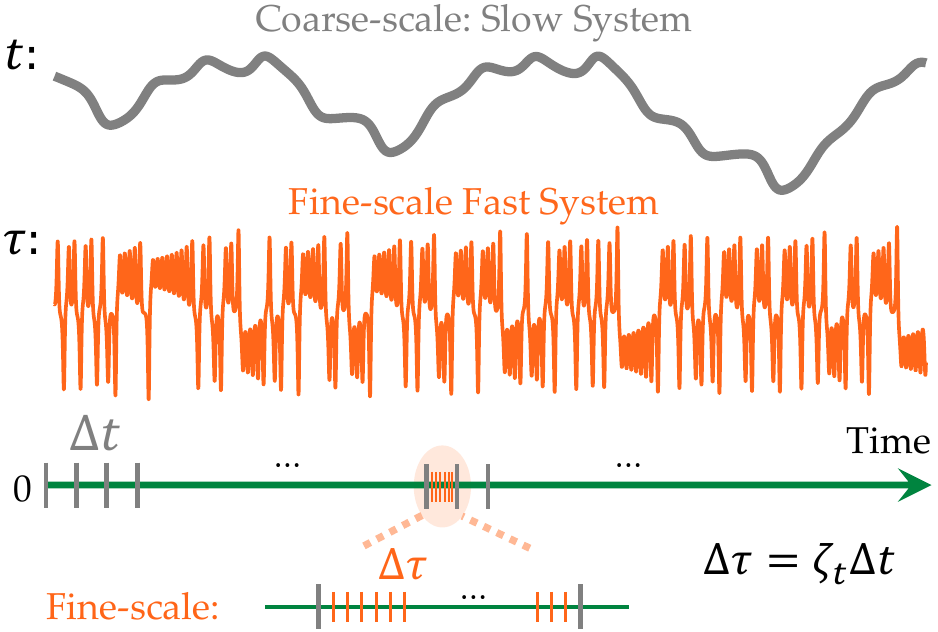} 
  
  \caption{Multi-scale sampling, where $\Delta \tau$ denotes the micro-scale time interval for fast dynamics, $\Delta t$ the macro-scale time interval for slow dynamics, and $\zeta_t$ the scale separation variable (typically $\zeta_t < 1$ or $\zeta_t \ll 1$).}
  \label{fig:sampling}
\end{figure}

Learning dynamical systems from data is a central challenge in scientific machine learning. PDE-governed physical systems are ubiquitous in disciplines such as biology, chemistry, and meteorology~\citep{anderson1995computational, darwish2016finite}. While traditional numerical solvers are reliable, direct numerical simulation (DNS) is often computationally prohibitive. It typically requires high spatial and temporal resolutions, long runtimes, and complete knowledge of the governing PDEs, model parameters, and initial/boundary conditions~\citep{ferziger2019computational, goc2021large}.

To alleviate the high computational cost of DNS, data-driven deep learning methods such as DeepONet~\cite{lu2021learning}, FNO~\cite{li2020fourier}, and DPOT~\cite{hao2024dpot, azizzadenesheli2024neural} aim to learn solution operators directly from data. These approaches bypass the need for explicit PDEs and enable fast inference by learning mappings between function spaces. During rollout, they operate with relatively large time steps, thereby reducing simulation time significantly compared to classical solvers. However, their success critically depends on the availability of large, high-quality datasets, which are often expensive to obtain via numerical simulations or physical experiments~\citep{kanov2015johns, li2024learning}.

To reduce data dependency, physics-informed learning has emerged as an effective paradigm. These methods embed physical knowledge into the training process either softly, by augmenting the loss function with PDE residuals, e.g., PINNs~\cite{raissi2019physics}, PINO~\cite{li2024physics}, MCNP~\cite{zhang2025monte}, or more strictly, by designing model architectures that obey physical rules through discrete approximations, e.g., PeRCNN~\cite{Rao_2023, rao2022discovering}, P$^2$C$^2$Net~\cite{WangEtAl2024NeurIPS}, LDSolver~\cite{yan2025ldsolver}, and TSM~\cite{sun2023neural}. While these approaches improve physical fidelity and reduce labeled data requirements, they still face challenges in long-term forecasting. Due to their reliance on local update schemes, they typically require small time steps for numerical stability, which can lead to cumulative errors and increased computational cost over long horizons.

Since the performance of neural solvers is closely tied to the quality and quantity of training data, the strategy employed during data acquisition plays a critical role. However, most existing works assume that data are sampled uniformly or continuously over time. This assumption does not hold in many real-world scenarios, where data collection is often constrained by factors such as high experimental costs, limited temporal resolution, and hardware limitations. In practical applications including mobile sensing, in situ physical experiments, and computational fluid dynamics simulations, data are frequently gathered in a sparse and non-uniform manner, often across multiple temporal scales. Such irregular sampling patterns introduce information gaps and temporal inconsistencies, which can significantly degrade the performance of learning-based models that rely on temporally regular inputs \cite{champion2019discovery}.

Among alternative sampling strategies, burst sampling has recently emerged as a promising approach. It captures multiple high-frequency samples within short time intervals to record fast-changing or transient dynamics, while alternating with low-frequency or missing data periods to conserve resources. This scheme has demonstrated the ability to capture both rapid and slow dynamics efficiently (see Figure~\ref{fig:sampling}). However, despite its growing practical relevance, burst-sampled, multi-scale data remains largely unexplored in the context of deep learning for spatiotemporal systems.

To address this underexplored yet practically important challenge, we propose a framework named Physics-Informed Multi-Scale Recurrent Learning (PIMRL). PIMRL is designed to model sparse, non-uniform, and burst-sampled spatiotemporal data by coupling latent-space macro-scale inference with micro-scale physical correction. The macro module operates in a learned latent space to perform efficient temporal reasoning and accelerate simulation, while the micro module enforces fine-grained physical consistency using pretrained PDE priors. To integrate information across time and resolution scales, we introduce a dedicated cross-scale message passing mechanism that fuses information between the two modules, ensuring stability and accuracy during long-term rollouts. Our contributions are summarized as follows:

\begin{itemize}
\item We propose a Physics-Informed Multi-Scale Recurrent Learning (PIMRL) framework that effectively leverages information from multi-scale, burst-sampled data to improve long-term spatiotemporal dynamics prediction.
\item We develop a cross-scale message passing mechanism that fuses physical information between macro and micro modules, ensuring stable latent state evolution and mitigating error accumulation during long rollouts. 
\item Our framework achieves optimal performance in effectively predicting the dynamics of diverse systems ranging from fluid flows to general physical phenomena, demonstrating scalability and generalizability under data-scarce, irregular sampling regimes.
\end{itemize}

\section{Related Work}
Accurate and efficient simulation of partial differential equations (PDEs) is essential for many scientific and engineering tasks, but traditional numerical methods are computationally costly. Although deep learning can accelerate simulations, these methods often require extensive physical knowledge or large datasets, resulting in poor performance when such resources are limited. In particular, current approaches struggle to model physical systems from sparse measurements without detailed physical priors.

\textbf{Direct Numerical Simulation.} 
Direct Numerical Simulation (DNS) is a computational method that solves the Navier–Stokes equations directly, capturing all spatial and temporal scales of turbulent flows without turbulence modeling \cite{anderson1995computational,darwish2016finite}. Although DNS yields highly accurate results, it is extremely computationally expensive. This immense computational cost and memory demand often limit DNS to relatively simple or small-scale problems, restricting its practical applicability. Furthermore, when the governing equations contain unknown parameters or missing physics, DNS may still struggle to provide reliable predictions. To overcome these challenges, PIMRL is specifically designed to efficiently predict outcomes by leveraging limited physical prior knowledge and data.

\textbf{Deep Learning Methods.} With the advancement of AI, its application to physical system simulation has become increasingly diverse and profound. For example, classical convolutional neural networks (CNNs) \cite{bar2019learning}, ResNet \cite{lu2018beyond}, graph neural networks (GNNs) \cite{sanchez2020learning,pfaff2020learning}, and transformer-based models \cite{wu2024transolver,hang2024unisolver,janny2023eagle,li2024scalable} have been widely adopted. Neural operators for learning mappings between function spaces have also seen rapid development, such as DeepONet \cite{lu2021learning}, FNO \cite{li2020fourier,rahman2022u,wen2022u}, and PIANO~\cite{zhang2024deciphering}.

\textbf{Physics-Informed Deep Learning Methods.}
To incorporate physical knowledge, researchers have developed various physics-informed approaches, such as PhyCRNet \cite{ren2022phycrnet}, PINN \cite{raissi2019physics}, PeSANet \cite{ijcai2025p862}, and PDE-Net \cite{long2018pde}. These methods embed physical laws into deep models to improve prediction accuracy. PeRCNN \cite{Rao_2023,rao2022discovering} employs a hard encoding of prior knowledge, enabling strong predictive and generalization capabilities even with limited data. However, most existing methods still suffer from error accumulation, which limits their long-term stability and accuracy. Notably, PeRCNN’s effective physical encoding strategy can be further leveraged in our proposed PIMRL framework.

\section{Methodology}

\begin{figure*}[t!]
   \vspace{-6pt}
    \centering
    \includegraphics[width=0.99\textwidth]{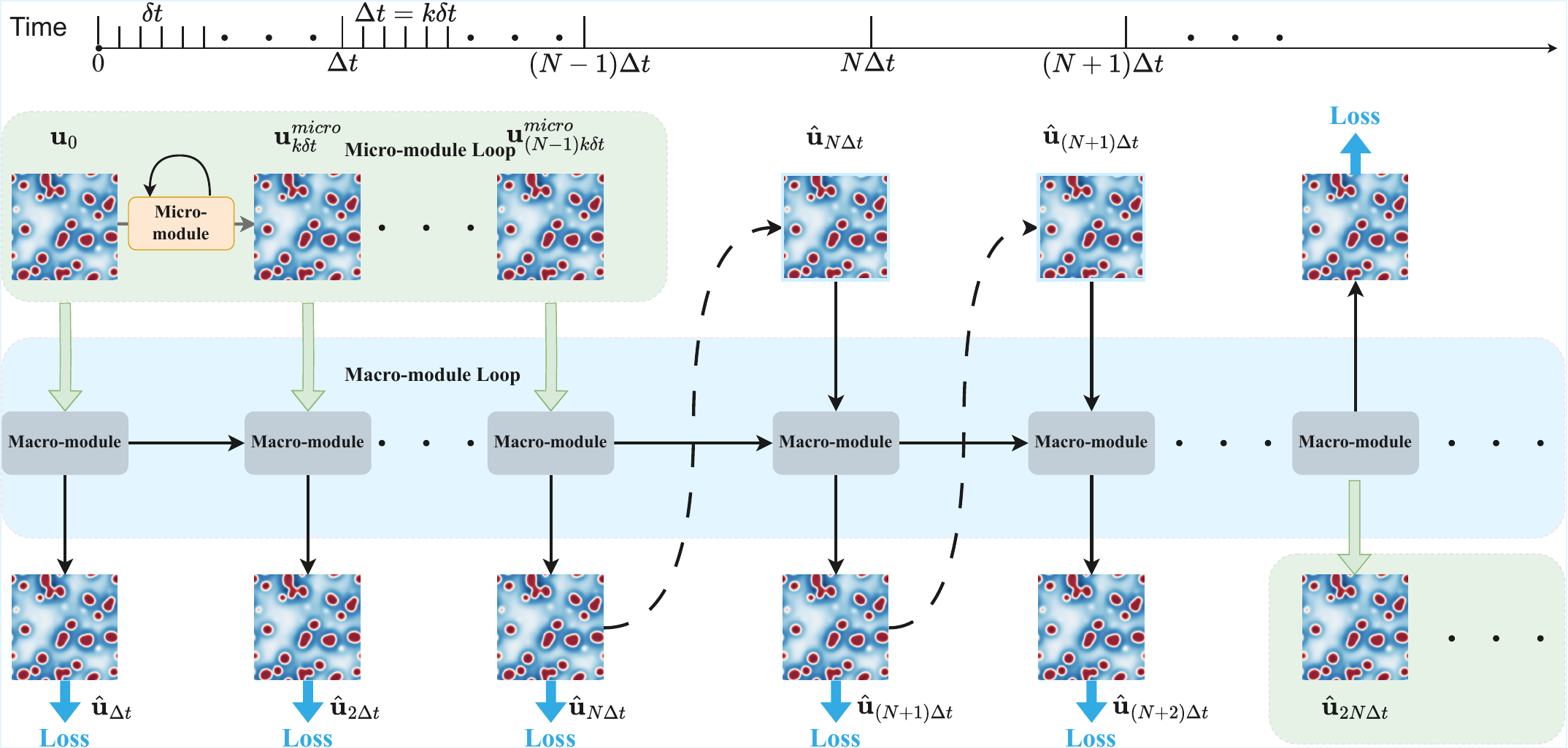}
    \vspace{0pt}
    \caption{ The overall framework architecture, which integrates physics-informed constraints with deep learning. The initial state of the system is denoted by $\mathbf{u}_0$. The state predicted by the micro-module after $k$ iterations, where each iteration occurs at intervals of $\delta t$, is represented by $\mathbf{u}^{micro}_{k\delta t}$. The predicted value of the physical state from PIMRL is denoted by $\hat{\mathbf{u}}$.}
    \label{mainmodel1}
    \vspace{-6pt}
\end{figure*}
\subsection{Partial Differential Equations Simulation Task }

Simulation tasks are closely related to PDEs, which fundamentally describe and model various physical phenomena, with the general form of a time-dependent PDE:
\begin{equation}
  \mathbf{u}_t = \mathcal{F}(t, x, \mathbf{u}, \nabla \mathbf{u}, \mathbf{u} \cdot \nabla \mathbf{u}, \nabla^2 \mathbf{u}, \ldots; \mu),
\end{equation}
where $\mathbf{u}(x, t)$ denotes the spatiotemporal solution field, $\mathbf{u}_t$ is the first-order time derivative, $\mathcal{F}(\cdot)$ is a (possibly nonlinear) operator, $\nabla$ represents the gradient (nabla) operator, $\nabla^2$ is the Laplacian, and $\mu$ denotes the set of PDE parameters.

In addition, the initial and boundary conditions (ICs and BCs) are specified as, for example, $I[\mathbf{u}](x, t = 0) = 0$ and $B[\mathbf{u}](x, t) = 0$, where $I[\cdot]$ and $B[\cdot]$ denote the operators for the initial and boundary conditions, respectively.

\subsection{Overview}
 
\begin{figure*}[t!]
    \centering
    \includegraphics[width=0.99\textwidth]{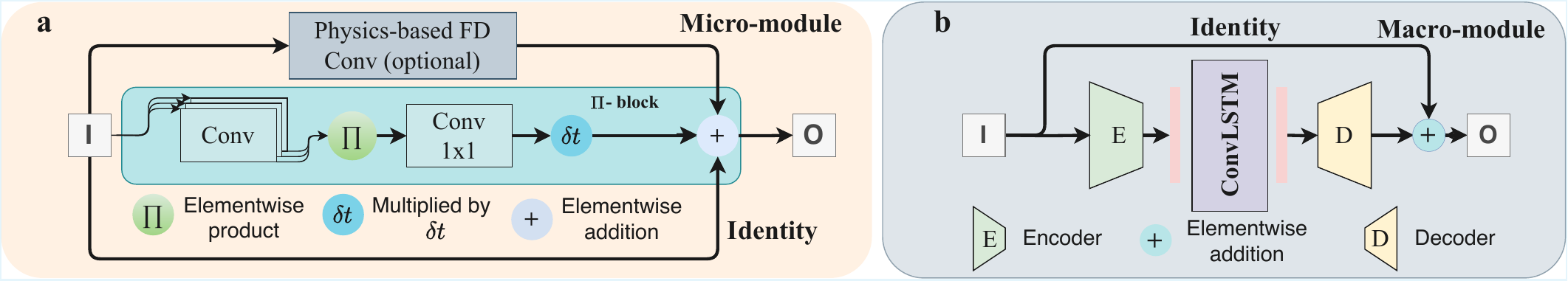}
    \vspace{-1pt}
    \caption{PIMRL includes two main modules: (a) the micro-module, designed to capture local features and small-scale dynamics; and (b) the macro-module, which captures long-term dependencies and global patterns using residual connections.}
    \label{mainmodel}
    \vspace{-6pt}
\end{figure*}

We aim to develop a model capable of predicting the evolution of spatiotemporal dynamics for nonlinear PDE systems over a long-term horizon, based on limited training data (e.g., a few trajectories generated from different initial conditions). As outlined in the introduction, we have designed the PIMRL to mitigate error accumulation and ensure that it effectively captures the intrinsic changes in the physical system, rather than overlooking significant physical information over extended time intervals. Furthermore, PIMRL is capable of efficiently utilizing multi-scale data.

As shown in Figure \ref{mainmodel1}, PIMRL consists of two main components: a macro-scale module and a micro-scale module. The macro-module performs fast, coarse-grained inference in the latent space with large time steps, accelerating prediction and reducing error accumulation. The micro-module conducts fine-grained reasoning and adjusts the latent historical states. The message-passing mechanism enables interaction between these modules: the micro-module provides physical knowledge and correction signals to the macro-module, while the macro-module passes updated information to guide future micro-module iterations.

PIMRL operates in a recursive manner, involving rollouts at the micro-module level, the macro-module level, and the overall framework level. Only the output from the macro-module contributes to the final output of the PIMRL framework and is used to compute the loss for training the entire framework, whereas the output from the micro-module refines and updates the latent historical states. The message-passing mechanism we proposed is described as follows:

\begin{itemize} 
\item Firstly, the micro-module rollout is a simple autoregressive process with time step $\delta t$, where the output at the previous time step serves as the input for the next step. 

\item When the micro-module is involved in the prediction, for every $k$ steps of micro-module with $\delta t$ like Equation \ref{eq2}, the final output of micro-module is passed to the macro-module, and at this point, the output from the macro-module serves as the output of the entire PIMRL model shown as Equation \ref{eq3}. When the micro-module is not involved in the prediction, the macro-module rollout is a simple autoregressive process with time step $\Delta t$.

\item Finally, there is a PIMRL loop that operates in conjunction with the macro-module rollout. After every $N-1$ rollout of the macro-module, the micro-module stops participating in the prediction, and the macro-module performs $N$ steps of autoregressive prediction on its own. This completes a total of $2N$ rollouts. Each output from the macro-module during these $2N$ rollouts serves as the output of PIMRL as depicted in Equation \ref{eq1}.
\end{itemize}

\subsection{Cross-scale Message Passing}

We conceptualize the interaction between the micro-scale module ($F_{\text{micro}}$) and the macro-scale module ($F_{\text{macro}}$) as a \textbf{cross-scale message passing} framework, illustrated in Figure \ref{mainmodel1}. The process is governed by the following operations:

First, during a data-rich burst, the micro-module generates a physically-informed ``message" by performing a high-fidelity rollout over $k$ fine steps. This message generation process is defined by the \textbf{Micro-module Loop}:
\begin{align}
\text{Micro-module Loop: } \mathbf{u}^{\text{micro}}_{t+k\delta t} = \underbrace{F_{\text{micro}}(...F_{\text{micro}}}_{\times k} (\mathbf{u}_t)) \label{eq2}
\end{align}

Next, this message, $\mathbf{u}^{\text{micro}}_{t+k\delta t}$, is passed to the macro-module, which consumes it to correct its own state. This state update is described by the \textbf{Macro-module Loop}:
\begin{align}
\text{Macro-module Loop: } \mathbf{\hat{u}}_{t+2k\delta t } = F_{\text{macro}}( \mathbf{u}^{\text{micro}}_{t+k\delta t} ) \label{eq3}
\end{align}

Finally, after being corrected by the message, the macro-module evolves autoregressively for efficient long-term forecasting. This complete, multi-step prediction cycle is summarized by the overall \textbf{PIMRL} process:
\begin{align}
\text{PIMRL: } \mathbf{\hat{u}}_{t+2N\Delta t }= \underbrace{F_{\text{macro}}(...F_{\text{macro}}}_{\times N} (\mathbf{\hat{u}}_{t+Nk\delta t})) \label{eq1}
\end{align}where the $\mathbf{u}$ denotes the physical state. The relationship between the micro-time step $\delta t$ and the macro-time step $\Delta t$ is given by $\Delta t = k \delta t$, where $k$ is an adjustable parameter.

\subsection{Micro-scale Module}
The micro-scale module is designed to learn underlying physical laws that govern the spatiotemporal dynamics from micro-scale data with small time stepping, where we adopt the PeRCNN model \cite{Rao_2023} with the architecture of $\Pi$-block shown in Figure \ref{mainmodel}(a). In a forward Euler scheme: $ \mathbf{u}_{(k+1)\delta t} =\hat{\mathcal{F}}(\mathbf{u}_{k\delta t}) \cdot \delta t +\mathbf{u}_{k\delta t},$ where $\delta t $ denotes that the module predicting in micro-scale time stepping. We can then approximate the $ \mathcal{F} $ by $\hat{\mathcal{F}}$ described as follows: 
\begin{equation}
\label{percnn}
    \hat{\mathcal{F}}(\mathbf{u}_{k\delta t}) = \sum^{N_c}_{c=1} W_c \cdot  \left[ \prod_{l=1}^{N_l}(K_{c,l} \star \hat{\mathbf{u}}_{k\delta t}+b_l)\right],
\end{equation}
where $N_c$ denotes the channel count, and $N_l$ the total number of parallel convolutional layers. The symbol $\star$ denotes the convolutional operation. For each layer $l$ and channel $c$, $K_{c,l}$ designates the specific filter weight, while $b_l$ stands for the bias term of that layer $l$. In the context of a $1 \times 1$ convolutional layer, $W_c$ denotes the weight assigned to the $c^{\text{th}}$ channel, with the bias term being omitted for the sake of simplicity and brevity. When a certain term in the governing PDE remains known (e.g., the Laplace operator $\nabla ^2 \mathbf{u}$), its discretization can be directly embedded in PeRCNN (called the physics-based Conv layer as shown in Figure \ref{mainmodel}(a)). The convolutional kernel in such a layer can be set according to the corresponding finite difference (FD) stencil. In essence, the physics-based Conv connection is constructed to incorporate known physical principles, whereas the $\Pi$-block is aimed at capturing the complementary unknown dynamics. The details of the physics-based FD Conv are provided in the Physical Filter part of Appendix A.

\subsection{Macro-scale Module} 
The macro-scale module serves as the long-term predictive module of PIMRL, specifically engineered to bridge the large temporal gaps inherent in burst-sampled data. Instead of operating in the computationally expensive physical space, it performs inference within a compact latent space. This design choice is motivated by the need for both computational efficiency and the expressive power to capture complex, long-range dependencies that are intractable for traditional, fine-grained solvers.

The module is implemented as a residual ConvLSTM autoencoder (Figure~\ref{mainmodel}; Appendix B), with components operating together within a message-passing framework. An encoder maps the input to a latent space, where the ConvLSTM cell propagates the system's state forward in time with large, efficient steps ($\Delta t$). Crucially, this latent-space evolution is not a pure black-box rollout. It is periodically corrected by physics-informed messages from the micro-module. This mechanism allows the macro-module to anchor its latent trajectory to physical reality, preventing the error accumulation and drift that plague purely data-driven recurrent models during long-term forecasting. By integrating the efficiency of latent-space dynamics with the physical fidelity of micro-scale corrections, our macro-scale module achieves a robust and accurate forecasting capability.

\begin{table*}[!ht]
\centering
\begin{tabular}{ l c c c c c c }
\toprule
Case & Numerical Methods & Spatial Grid & Macro Step& Micro Step & Training Trajectories & Test Trajectories \\
\midrule
KdV & FVM & 256 & 0.01s& 0.01s & 5 & 2 \\
Burgers & FDM & $128^2$ & 0.001s& 0.01s & 13 & 3 \\
FN & FDM & $128^2$ & 0.5s& 0.01s & 5 & 3 \\
2D GS & FDM & $128^2$ & 0.002s & 0.01s& 2 & 3 \\
3D GS & FDM &$96^3$ & 0.25s& 0.01s & 3 & 2 \\
\hline
\end{tabular}
\caption{Summary of experimental settings for different cases. The 3D GS case is downsampled from $96^3$ to $48^3$ during training.}
\label{tab:experimental_settings}
\vspace{0pt}
\end{table*}

\section{Experiments}
To validate the effectiveness of our proposed PIMRL framework, we conducted extensive experiments on a diverse set of fluid dynamics and reaction-diffusion systems. 

\begin{figure*} [!h]
  \centering
  \includegraphics[width=0.84\textwidth]{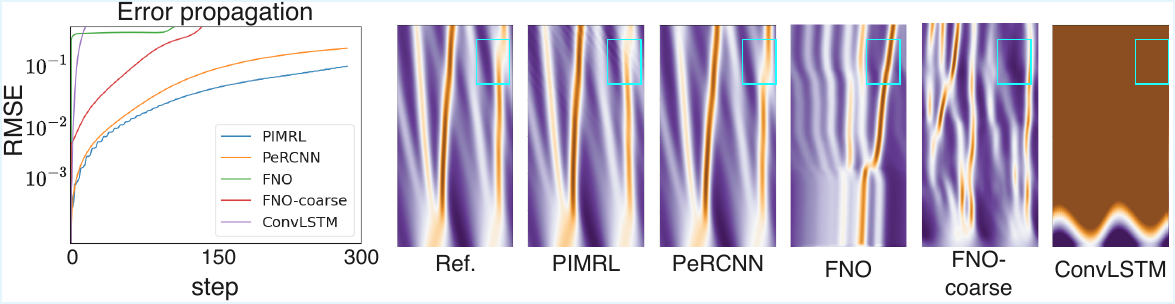} 
  \vspace{-6pt}
\caption{Error propagation curves and final prediction plots for the KdV case, comparing the PIMRL and baseline models.}
  \label{fig:1d}
  \vspace{-3pt}
\end{figure*}

\begin{figure*}[!h]
    \centering
    \includegraphics[width=0.86\textwidth]{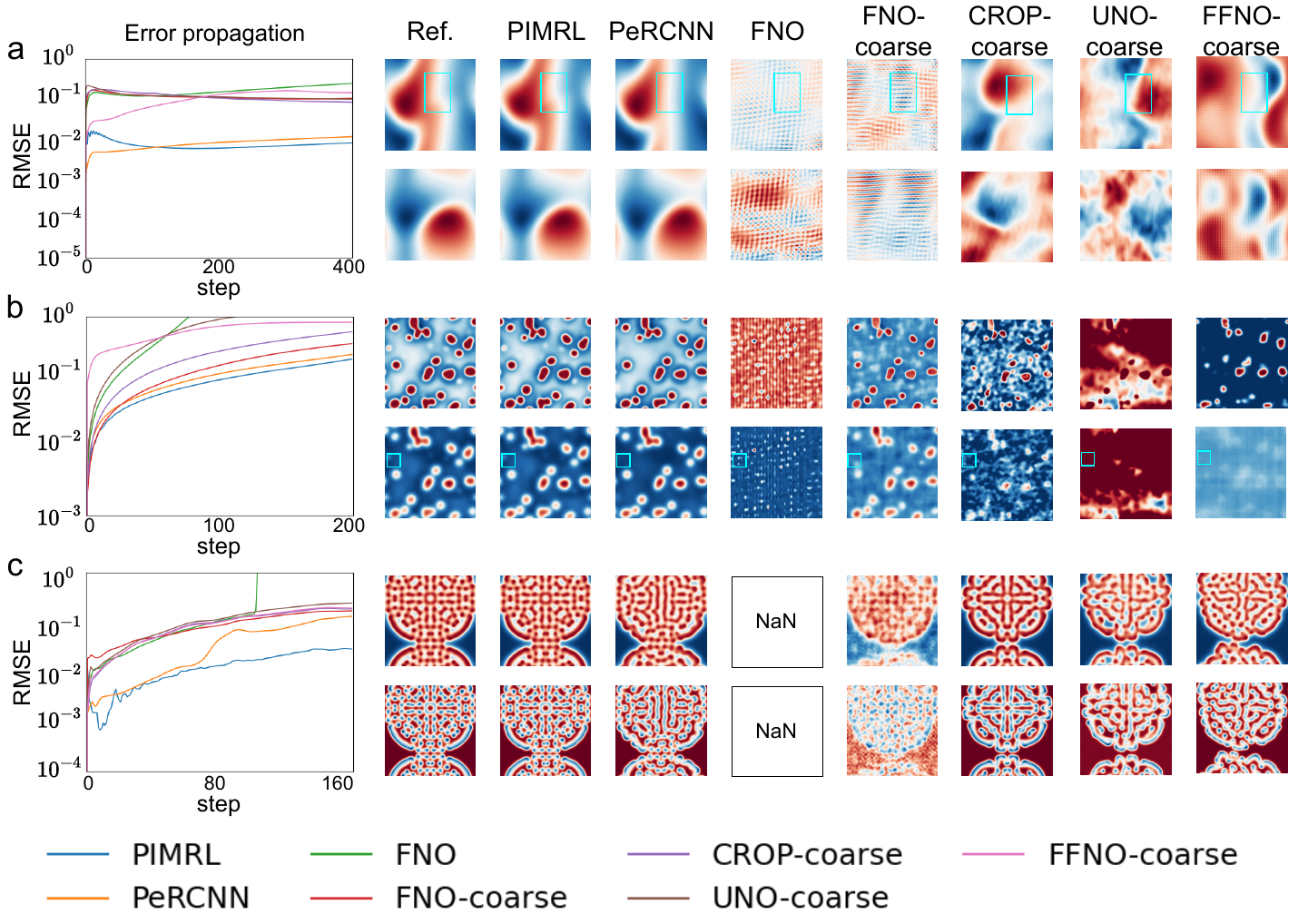}
    \caption{Figures (a)–(c) summarize qualitative comparisons with baselines on Burgers, FN, and 2D GS, showing error propagation and final predictions.}
    \label{results}
\end{figure*}

\textbf{Datasets.}
We conducted experiments on five datasets, including Korteweg-de Vries (KdV), Burgers, FitzHugh-Nagumo (FN), 2D Gray-Scott (2D GS), and 3D Gray-Scott (3D GS). 
The IC for the KdV equation is created by summing multiple sine waves with random amplitudes, phases, and frequencies, resulting in a complex waveform. ICs for the Burgers' equation are generated randomly according to a Gaussian distribution. The FN equation is initialized with random Gaussian noise for a warm-up period, after which time sequences are extracted to form the dataset. The GS equation starts the reaction from random initial positions and then diffuses. In these cases, except for the KdV case, which was solved using the Finite Volume Method (FVM), the rest were solved using the Finite Difference Method (FDM).

Additionally, we have two sets of data with different time scales originating from the same ICs. The micro-scale data $\mathbf{U}^{\text{micro}} = \{\mathbf{u}_0, \mathbf{u}_{\delta t}, \mathbf{u}_{2\delta t}  \cdot \cdot \cdot   \} \in \mathbf{R}^{\text{micro}}$ is characterized by short and scattered continuous time intervals, while the macro-scale data $\mathbf{U}^{\text{macro}} = \{\mathbf{u}_0, \mathbf{u}_{\Delta t}, \mathbf{u}_{2\Delta t}  \cdot \cdot \cdot  \mathbf{u}_{T_\text{end}} \} \in \mathbf{R}^{\text{macro}}$ exhibits persistent continuity until the end. The validation and test sets are established based on different ICs but with the same parameters, making it more challenging than extrapolation under the same ICs.

\textbf{Baseline Models.} To validate the effectiveness of the proposed PIMRL framework, we introduced several baseline models. Firstly, we considered the widely recognized high-performing data-driven model FNO \cite{li2020fourier}, which has been trained on datasets with two different time intervals, denoted as FNO (trained on fine-scale data with small time steps) and FNO-coarse (trained on coarse-scale data with large time steps). Secondly, we included the PeRCNN model \cite{Rao_2023}, which embeds physical knowledge in a hard way. Due to constraints on the time stepping of the model, PeRCNN is trained on datasets with small time steps. Lastly, we incorporated CROP~\cite{gao2025discretizationinvariance}(CROP-coarse trained on coarse-scale data with large time steps), a method proposed to address problems with generalization and discretization mismatch errors in existing neural operators across different data resolutions. Since the original CROP paper did not include 1D and 3D cases, we employ AE-ConvLSTM \cite{vlachas2022multiscale} as a substitute. The details are shown in Appendix C.

\textbf{Evaluation Metrics.} To comprehensively evaluate the performance of our model, we adopted metrics: Root Mean Square Error (RMSE), Mean Absolute Error (MAE), and High Correction Time (HCT). RMSE is calculated on the macro-scale data to facilitate comparisons between models operating at different granularities. MAE provides a measure of the average absolute difference between the predicted and actual values. HCT evaluates the time it takes for the model to correct its predictions to a high level of accuracy. Detailed formulas for these metrics are provided in the Evaluation Metrics part of Appendix E.

\begin{table}[!ht]
\centering

\resizebox{0.472\textwidth}{!}
{\begin{tabular}{ccccc}
\toprule
Case & Model & RMSE $\downarrow$& MAE $\downarrow$& HCT (s) $\uparrow$\\ \midrule
\multirow{6}{*}{KdV} 
& ConvLSTM & 5.8507 & 7.6036 & 9.6\\ 
& FNO & 0.4891 & 0.3300 & 0.45\\ 
& FNO-coarse & 0.5461 & 0.4167 & 7.8 \\ 
& PeRCNN & \underline{0.0942}& \underline{0.0941} & \underline{30} \\ 
\cmidrule{2-5}
& PIMRL(Ours) & $\mathbf{0.0457}$  & $\mathbf{0.0607}$  & $\mathbf{46.2}$ \\ 
& Promotion  & 51.5\% & 35.5\% & 54.0\%   \\  
\midrule
\multirow{6}{*}{Burgers} 
& CROP-coarse & 0.1103 & 0.0887 & 0.064 \\ 
& FNO & 0.1561 & 0.1301 & 0.104 \\ 
& FNO-coarse & 0.1094 & 0.0879 & 0.064 \\ 
& PeRCNN & \underline{0.0075}& \underline{0.0058} & \underline{3.216} \\ 
\cmidrule{2-5}
& PIMRL(Ours) & $\mathbf{0.0068}$  & $\mathbf{0.0049}$  & $\mathbf{3.216}$ \\ 
& Promotion  & 9.3\% & 15.6\% &0\%$^{*}$  \\      
\midrule
\multirow{6}{*}{FN} 
& CROP-coarse  & 0.3950 & 0.2734 & 3.03 \\ 
& FNO & 937 & 2393980 & 1.65 \\ 
& FNO-coarse & 0.1878 & 0.1643 & 4.98 \\ 
& PeRCNN & \underline{0.1591}& \underline{0.1139} & \underline{6.99} \\ 
\cmidrule{2-5}
& PIMRL(Ours) & $\mathbf{0.1349}$  & $\mathbf{0.0990}$  & $\mathbf{7.74}$ \\ 
& Promotion  & 15.2\% & 13.1\% & 10.7\%  \\     
\midrule
\multirow{6}{*}{2D GS} 
& CROP-coarse  & 0.1027 & 0.0579 & 1260 \\ 
& FNO & NaN & NaN & 810 \\ 
& FNO-coarse & 0.0884 & 0.0629 & 1335 \\ 
& PeRCNN & \underline{0.0455}& \underline{0.0268} & \underline{1379.5} \\ 
\cmidrule{2-5}
& PIMRL(Ours) & $\mathbf{0.0133}$  & $\mathbf{0.0072}$  & $\mathbf{1965^{*}}$ \\ 
& Promotion  & 70.8\% & 73.1\% & 42.5\%   \\     
\midrule
\multirow{6}{*}{3D GS} 
& ConvLSTM & 0.2081 & 0.2009 & 56.5 \\ 
& FNO & 0.2798 & 0.1950 & 112.5 \\ 
& FNO-coarse & 0.1042 & 0.0611 & 360 \\ 
& PeRCNN & \underline{0.0532}& \underline{0.0977} & \underline{510} \\ 
\cmidrule{2-5}
& PIMRL(Ours) & $\mathbf{0.0381}$  & $\mathbf{0.0190}$  & $\mathbf{731.25}$ \\ 
& Promotion  & 28.4\% & 80.6\% & 43.4\%   \\      
\bottomrule
\vspace{-10pt}
\end{tabular}}

\caption{Quantitative results of our model and baselines, where $*$ denotes that the prediction lasts to the end.}
\label{tab:model_performance}
\end{table}

\subsection{Main Results}\label{main results}
Figure \ref{results} compares our framework with baseline models across cases using box plots of predicted quantities. Table \ref{tab:model_performance} reports the corresponding quantitative metrics.

\textbf{1D KdV Equation.} The KdV equation governs nonlinear wave evolution. Notably, significant discrepancies between PeRCNN predictions and ground truth emerge in the blue box of Figure \ref{fig:1d}, whereas other baselines exhibit poor initial performance attributable to KdV complexity. Conversely, PIMRL demonstrates substantial advantages in predictive accuracy over baseline models, achieving 30\%–50\% improvements in evaluation metrics (Table \ref{tab:model_performance}) and thereby highlighting significant field advancement.

\textbf{2D Burgers Equation.} Figure \ref{results}(a) demonstrates that only PeRCNN and our framework attain satisfactory post-training results on micro-scale data, while other baselines fail to capture physical changes. In the blue-boxed regions, PeRCNN shows significant long-term forecast errors compared to PIMRL's sustained accuracy. Corresponding quantitative improvements in RMSE and MAE are documented in Table \ref{tab:model_performance}, with PeRCNN's divergence further confirmed at the final time step.

\textbf{2D FitzHugh-Nagumo Equation.}
In Figure \ref{results}(b), the purely data-driven method fails to achieve its original performance on this relatively small dataset. Quantitatively, our model outperforms the best existing model by at least 10\%. Additionally, Table \ref{tab:model_performance} shows that while the RMSE and other metric curves of our model and PeRCNN are similar, our model consistently achieves better results.

\textbf{2D Gray-Scott Equation.} As shown in the left part of Figure \ref{results}(c), only PIMRL effectively carried out long-term predictions, showcasing a clear demonstration of the cumulative errors of PeRCNN in this case. In this case, Table \ref{tab:model_performance} further validates the superior performance of PIMRL with both RMSE and MAE showing an improvement of over 70\%, and the HCT similarity of PIMRL at 0.99 at the final prediction step. This highlights the effectiveness of PIMRL in improving predictive accuracy and consistency throughout forecasting, demonstrating its potential for tackling complex challenges.

\begin{figure} [t!]
  \centering
  \includegraphics[width=0.45\textwidth]{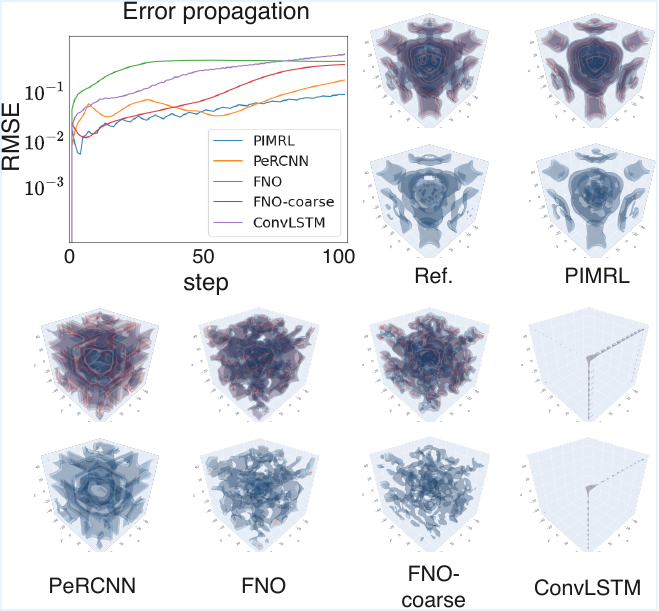} 
  \caption{Error propagation curves and final prediction plots for the 3D GS, comparing the PIMRL and baseline models.}
  \label{fig:3d}
\end{figure}

\textbf{3D Gray-Scott Equation.}
As shown on the right side of Figure \ref{fig:3d}, the predictions under our framework closely align with the ground truth. Through box plots and error evolution graphs, it is evident that PeRCNN, as the best-performing model among the baselines in our study, outperformed the other models lacking physical knowledge embeddings, especially when compared to FNO trained on the same dataset. In quantitative analysis of Table \ref{tab:model_performance}, the improvements are also significant, not only in the overall evaluation metrics of RMSE and MAE in 28.4\% and 80.6\% but also in the substantial growth of the HCT in 43.4\%.

\subsection{Ablation Study}

We designed five novel models and report them in Table \ref{ablation}. 

\begin{table}[t!]
    \centering
    \footnotesize
    \begin{tabularx}{\linewidth}{lc}
        \toprule
        Model & RMSE \\
        \midrule
        PIMRL w/o Connect & 0.1975 \\
        FNO-MRL & 0.7854 \\
        PIMRL w/o Pretraining & 0.2599 \\
        PIMRL w/o Physics-based FD Conv & 0.1738 \\
        PIMRL & 0.1349 \\
        \bottomrule
    \end{tabularx}
    \caption{Results for ablation study in FN-case.}
    \label{ablation}
    \vspace{-6pt}
\end{table}

(1) The ablation study with the ``PIMRL w/o Connect'', where the connections between the micro-module and the macro-module are removed, is designed to demonstrate the effectiveness of the PIMRL framework's structural design. This experiment, which leaves only the serial structure of the two modules, shows that the connections within the PIMRL framework are essential. (2) The ``FNO-MRL'' replaces the micro-scale module containing physical information, PeRCNN, with the data-driven model FNO, aiming to validate the efficacy of the physical embedding in the micro-scale module within our framework.
(3) ``PIMRL w/o Pretraining'' provides a perspective on the training method by eliminating the pretraining step for the micro module. In this ablation study, the absence of pretraining leads to inferior performance compared to the PIMRL. This demonstrates that directly introducing large time intervals for training can deprive the micro module of the opportunity to learn fine-grained changes, similar to why PeRCNN cannot be directly used with large time intervals. (4) ``PIMRL w/o Physics-based FD Conv'' indicates the removal of the Physics-based FD Convolution. This ablation study emphasizes the effectiveness of the Physics-based FD Conv by showing the performance degradation when it is omitted. (5) ``PeRCNN w/o Physics-based FD Conv'' is a version of PeRCNN without the Physics-based Finite Difference (FD) Convolution.

\subsection{Addition Results}

To comprehensively evaluate PIMRL, we conducted a series of supplementary experiments, the details of which are provided in Appendix G. In particular, we assessed computational efficiency by comparing runtime, parameter count, and memory usage with other physics-embedded methods, and examined the model’s scalability and practicality in various scenarios. Furthermore, we investigated the robustness of PIMRL under different training set sizes and analyzed the reliability of its predictions using error bar analyses. These additional experiments collectively demonstrate the promising efficiency, scalability, and robustness of PIMRL.

\section{Conclusion}

In this paper, we introduced the Physics-Informed Multi-Scale Recurrent Learning (PIMRL) framework, the first, to our knowledge, specifically designed to predict the evolution of spatiotemporal systems from multi-scale, burst-sampled data. Its novel architecture synergistically combines a macro-scale module for efficient long-term forecasting with a physics-informed micro-scale module for periodic correction, a design that effectively mitigates cumulative error while capturing underlying physical laws from extremely limited observations. Rigorous evaluations confirmed its substantial performance gains over state-of-the-art baselines, including a reduction in RMSE of up to 70\% on the 2D Gray-Scott dataset, thereby validating the effectiveness of this multi-scale, physics-informed approach. 

While we acknowledge the framework's current limitations, such as the yet unexplored sensitivity to the duration of data gaps and the presence of input noise, as detailed in Appendix H, we view them as clear and important avenues for future research. Going forward, we will improve the macro-module’s efficiency and integrate advanced super-resolution to better exploit the model’s potential, enhance generalization, and widen its impact in scientific computing.

\section{Acknowledgments}
The work is supported by the Beijing Natural Science Foundation (No. 1232009) and the National Natural Science Foundation of China (No. 62276269 and No. 62506367). R.Z. would like to acknowledge the supported by the China Postdoctoral Science Foundation under Grant Number 2025M771582 and the Postdoctoral Fellowship Program of CPSF under Grant Number GZB20250408. Additional experimental details and full appendices are available in the arXiv version of this paper at \url{https://arxiv.org/abs/2503.10253}.

\bibliography{aaai2026}

\section{Impact statement}
The paper endeavors to devise a adaptable framework that expedites simulations and predictive analyses of physical systems by utilizing multi-scale temporal data. This framework harmoniously integrates data-driven methodologies with physics-informed principles, striking a delicate balance between empirical insights and theoretical underpinnings in its application. The framework can be widely applied in various research fields including material science, turbulent flow prediction, chemical engineering, and so forth. Our research is exclusively conducted for the pursuit of scientific objectives and does not entail any potential ethical concerns or risks.

\section{A: Physical Filter}
\label{sectionb}
In the one-dimensional problem KdV, the original paper of PeRCNN did not provide the corresponding model design. Following their concept, we present the corresponding physical filter.
\begin{equation}
    f_{\text{xxx}} = \frac{-f_{i-2} + 2f_{i-1} + 0f_i - 2f_{i+1} + f_{i+2}}{2h^3}
\end{equation}

As shown in the Figure \ref{filter4kdv}, we designed a Physical-Filter to represent $\frac{\partial^3 u}{\partial x^3}$ in the KdV equation. This approach leverages the inherent physical properties of the system to accurately model the third-order spatial derivative, thereby enhancing the accuracy and efficiency of the numerical solution. Among the parameters, $h$ indicates the $\Delta x$ in the cases. The Boundary Padding is an approach to adapt to periodic boundary conditions by replacing the original zero padding with a periodic boundary padding.

\section{B: Implementation Details}
\subsection{Overview}
\label{sectioncov}
The overview of the PIMRL framework, which includes a pretraining stage using micro-scale data for physics-informed Learning and the utilization of a micro-module, informed by learned physics knowledge, to correct the macro-module during training on macro-scale data.
\begin{figure*}[!t]

    \centering
    \includegraphics[width=0.9\textwidth]{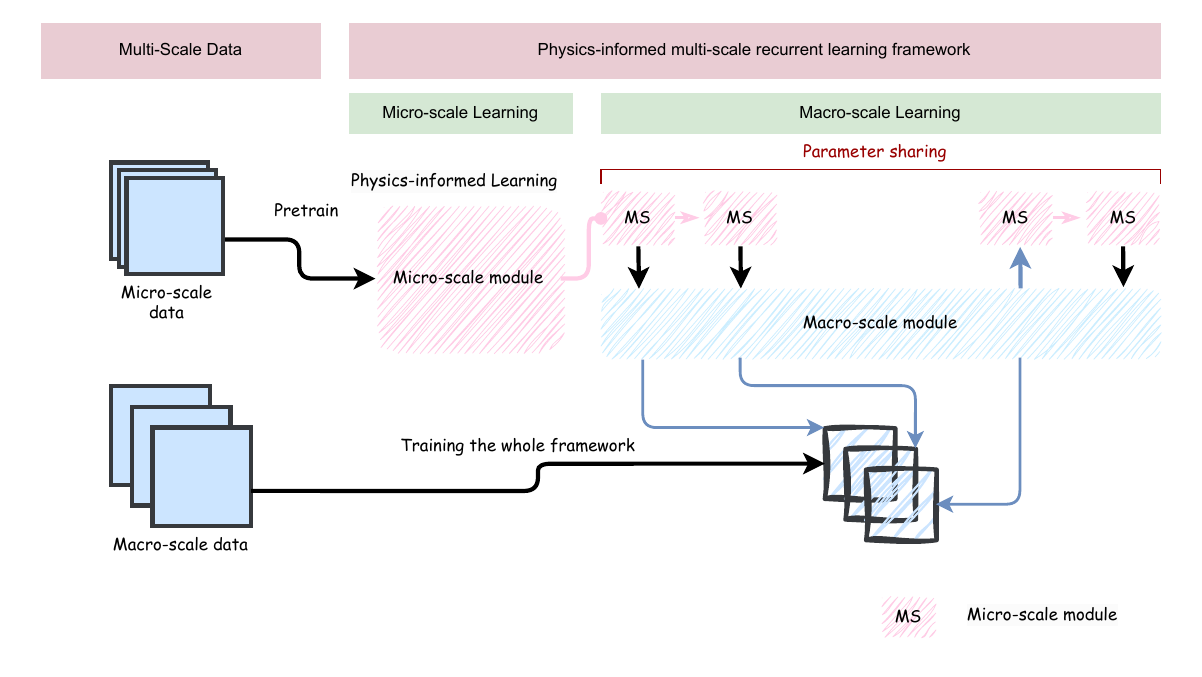}
    \vspace{-10pt}
    \caption{{ Overview of the PIMRL framework, which includes a pretraining stage using micro-scale data for physics-informed Learning and the utilization of a micro-module, informed by learned physics knowledge, to correct the macro-module during training on macro-scale data.}}
    \label{overview}
    \vspace{-10pt}
\end{figure*}

\label{sectionc}
In the main body of this paper, we have elaborated on the architectures at the micro and macro scales. Within the macro-scale module, there are components including an encoder, a decoder, and a Residual Long Short-Term Memory (ResidualLSTM). The following section will provide a detailed exposition of their configuration specifics.
\subsection{Encoder and Decoder}
\label{sectionc1}
In our framework, the autoencoder is employed not for the purpose of minimizing reconstruction loss. Instead, the encoder is utilized to extract features, while the decoder serves to project the output of ConvLSTM into the physical space as residuals. The primary goal of the autoencoder in this context is to map an input to a low-dimensional latent space, and subsequently decode it to the original dimension at the output, facilitating the feature extraction and residual projection processes in the framework.

\subsection{ResidualLSTM}
\label{sectionc2}
\begin{figure}[h]

    \centering
    \includegraphics[width=0.5\textwidth]{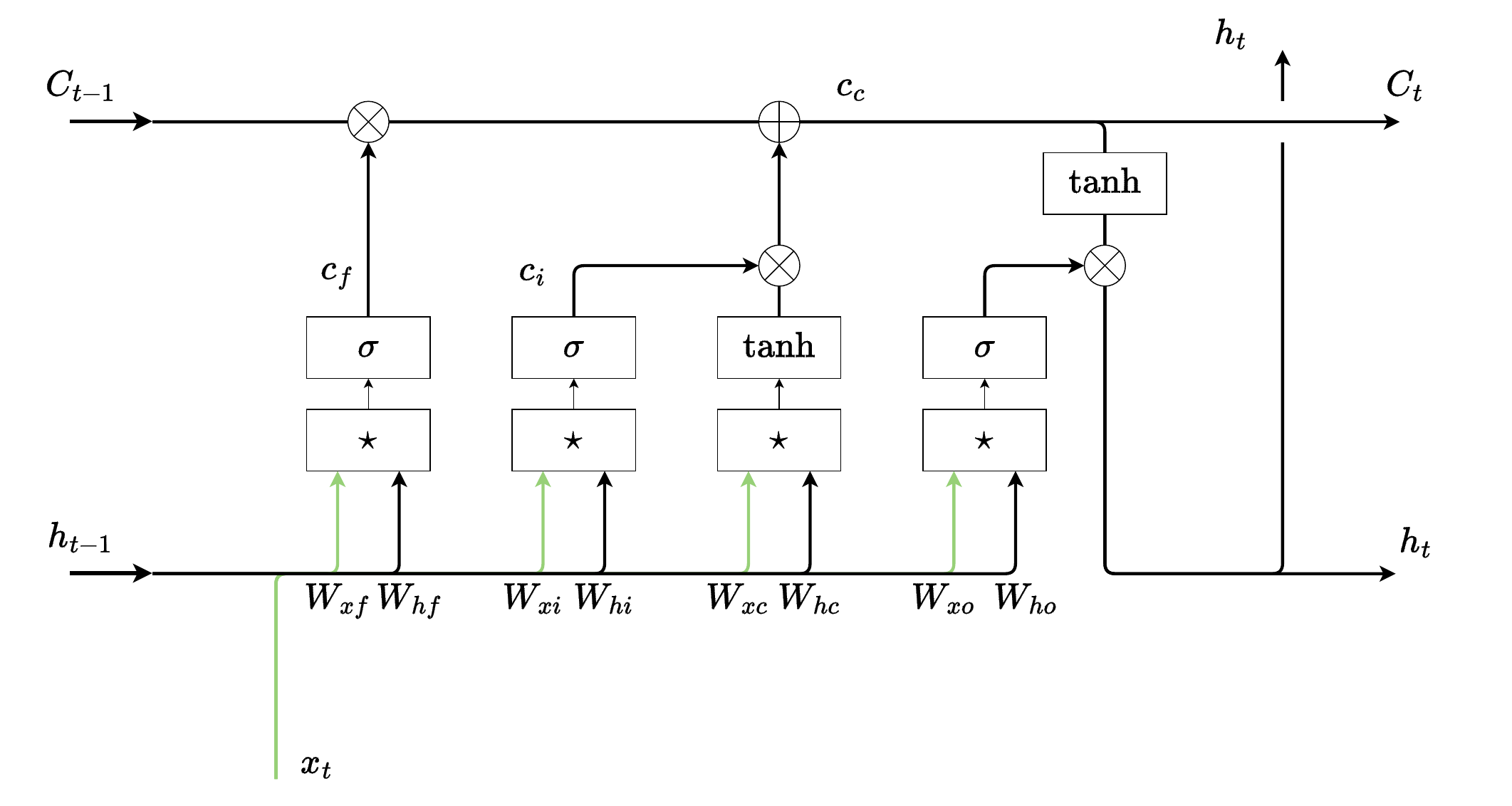}
    \caption{ConvLSTMCell}
    \label{modelcvlstmcell}
\end{figure}

The structure of ResidualLSTM has been clearly illustrated in the main text. Here, we will elucidate the architecture of the ConvLSTMcell as the Figure \ref{modelcvlstmcell}.

\begin{equation}
\label{cell}
\begin{aligned}
    c_i &= \sigma(\text{W}_{xi}(x_{t}) + \text{W}_{hi}(h_{t-1})) \\
c_f &= \sigma(\text{W}_{xf}(x_{t}) + \text{W}_{hf}(h_{t-1})) \\
c_c &= c_f \cdot c + c_i \cdot \tanh(\text{W}_{xc}(x_{t}) + \text{W}_{hc}(h_{t-1})) \\
C_t &= \sigma(\text{W}_{xo}(x_{t}) + \text{W}_{ho}(h_{t-1})) \\
h_t &= c_o \cdot \tanh(c_c)
\end{aligned}
\end{equation}

The set of equations \ref{cell} presented above outlines the operations of a ConvLSTM cell. The equations involve computations of various gates and states within the cell, including input gate $c_i$, forget gate $c_f$, cell state $c_c$, output $C_t$, and the updated hidden state $h_t$. These equations govern the flow of information and transformations within the ConvLSTM cell, enabling the model to process spatiotemporal data efficiently by considering spatial dimensions in the calculations. The parameters $\text{W}{xi}$, $\text{W}{hi}$, $\text{W}{xf}$, $\text{W}{hf}$, $\text{W}{xo}$, and $\text{W}{ho}$, depicted in Figure \ref{modelcvlstmcell}, manage information from inputs ($x_{t}$) and history ($h_{t-1}$,$C_{t-1}$) in a convolutional manner.
 
\begin{table*}[ht]
\centering

\begin{tabular}{lcccc}
\toprule
\textbf{Metrics} & \textbf{U-NO} & \textbf{F-FNO} & \textbf{MWT} & \textbf{FNO} \\
\midrule
RMSE & 0.3675 & 0.2280 & 0.3494 & 0.1878 \\
MAE & 0.1465 & 0.1350 & 0.2228 & 0.1634 \\
\bottomrule
\end{tabular}
\caption{The Result of U-NO, F-FNO, MWT and FNO in the FN Case.}
\label{tab:fn_results}
\end{table*}

\section{C: Baseline Models}

In order to compare and evaluate the performance of our proposed method, we have trained multiple state-of-the-art (SOTA) baseline models as well as classical models, and compared them with our model. The introductions to each baseline model are presented below, while the training details are outlined in subsequent sections \ref{sectione}.

\textbf{Fourier Neural Operator (FNO) \cite{li2020fourier}.} FNO is a method that combines Fourier transforms with neural networks. This approach comprises two main components. The first component involves performing Fourier transforms on the system state quantities, learning certain information in the frequency domain, and then applying an inverse transform. The second component utilizes convolutions to process the system state quantities, complementing the information not captured during the frequency domain learning. The combination of these two components serves as the final result. We make two FNO models sharing the same architecture, which training in micro and macro scale datasets respectively and inferring in micro and macro time intervals. In the same model, we conducted training separately for two types of data, resulting in FNO and FNO-coarse.

\textbf{ConvLSTM \cite{vlachas2022multiscale}.} ConvLSTM is a specialized neural network architecture that combines convolutional and LSTM layers to effectively model spatial and temporal dependencies in sequential data. We use AE-ConvLSTM as our base model.

\textbf{PeRCNN \cite{Rao_2023}.} PeRCNN represents a physics-informed learning methodology, embedding physical laws directly into the neural network architecture. It incorporates multiple parallel convolutional neural networks (CNNs), leveraging the simulation of polynomial
equations through feature map multiplication. By doing so, PeRCNN augments the model's extrapolation and generalization capabilities.

\section{D: Training Details}
\label{sectione}
All experiments were conducted on a single 80GB Nvidia A100 GPU, using an Intel(R) Xeon(R) Platinum 8380 CPU (2.30GHz, 64 cores). We only give some of the changed parameters here, and the other hyperparameters remain the same as the original text.

\textbf{PIMRL.} The micro-module has been trained like PeRCNN  in the same way. The architecture of the whole PIMRL model, illustrated in the paper, utilizes the Adam optimizer with a learning rate of $5 \times 10^{-3}$. The model have different parameters in different cases. 

Additionally, we implement the StepLR scheduler to adjust the learning rate by a factor of 0.98 every 200 epochs. The pretraining details is same to the baseline model PeRCNN.

\textbf{FNO and FNO-coarse.} The network structure of FNO remains largely in line with the original study, with the primary adjustment being the adoption of an autoregressive training method for this model. We employ the Adam optimizer with a learning rate of $1 \times 10^{-3}$. 

\textbf{ConvLSTM.} For ConvLSTM, we implement the ConvLSTM architecture like the macro-scale module of PIMRL. The StepLR scheduler is utilized with a step size of 200 and a gamma value of 0.98. The optimizer of choice is Adam, featuring a learning rate set at $1 \times 10^{-3}$. 

\textbf{PeRCNN.} For PeRCNN, the training is different from the original paper since the micro-scale data only get a few pairs of continuous data. It is impossible to train PeRCNN in 400 to 800 steps at the same time. We employ the Adam optimizer with learning rate of $1 \times 10^{-3}$ and the StepLR scheduler to adjust the learning rate by a factor of 0.98 per 200 epochs.

\section{E: Evaluation Metrics}
\label{sectionf}
We have adopted some classical evaluation metrics such as RMSE, MAE and HCT. Root Mean Square Error (RMSE) quantifies the average error magnitude between estimated and actual values, serving as a gauge of the model's precision. Conversely, Mean Absolute Error (MAE) assesses the average absolute disparity between anticipated and observed values, denoting the true scale of discrepancies.

The definitions of these metrics are as follows:

\begin{equation}
\label{metric}
\begin{aligned}
\text{RMSE (Root Mean Square Error):} & \quad \sqrt{\frac{1}{n} \sum_{i=1}^{n} (y_i - \hat{y}_i)^2} \\
\text{MAE (Mean Absolute Error):} & \quad \frac{1}{n} \sum_{i=1}^{n} |y_i - \hat{y}_i|\\
\text{HCT (High Correction Time):}& \sum_{i=1}^{N} \Delta t \cdot 1(\text{PCC}(y_i, \tilde{y}_i) > 0.8)
\end{aligned}
\end{equation}
In the above equations\ref{metric}, $n$ represents the number of trajectories, $y_i$ represents the true value, and $ \hat{y}_i$ represents the predicted value of the model. The $\text{PCC}$ is the Pearson correlation coefficient, which is a statistical metric used to measure the linear correlation between two variables.

\section{F: Dataset Informations}
The IC for the Kdv equation is created by summing multiple sine waves with random amplitudes, phases, and frequencies, resulting in a complex waveform. Initial conditions for the Burgers' equation are generated randomly according to a Gaussian distribution. The FN equation is initialized with random Gaussian noise for a warm-up period, after which time sequences are extracted to form the dataset. The GS equation starts the reaction from random initial positions then diffuses.

\begin{table*}[!ht]
 
\centering

\begin{tabular}{lccc}
\hline
Model      & Running Time & Parameter Size & GPU Memory \\ \hline
PIMRL      & 9 s          & 3.33 M         & 1728 M     \\  
U-NO       & 7 s          & 15.29 M        & 2320 M    \\  
MWT        & 12 s         & 0.09 M         & 1732 M    \\ 
FNO        & 2 s          & 8.39 M         & 1580 M    \\  
ConvLSTM   & 5 s          & 3.32 M         & 1708 M    \\ \hline
\end{tabular}
\caption{Running Time, Parameter Size, and GPU Memory for PIMRL, U-NO, MWT, FNO, and ConvLSTM}
\label{tab:performance_comparison}
\end{table*}

\begin{table*}[ht]
\centering

\begin{tabular}{ l c c c c c }
\toprule
Case & Numerical Methods & Spatial Grid & Time Grid & Training Trajectories & Test Trajectories \\
\midrule
Kdv & FVM & 256 & 0.01s & 5 & 2 \\
Burgers & FDM & $128^2$ & 0.001 & 13 & 3 \\
FN & FDM & $128^2$ & 0.5 & 5 & 3 \\
2DGS & FDM & $128^2$ & 0.002 & 2 & 3 \\
3DGS & FDM &$96^3$ & 0.25 & 3 & 2 \\
\hline
\end{tabular}
\caption{Summary of experimental settings for different cases.(The 3D GS case is downsampled from $96^3$ to $48^3$ during training)}
\label{tab:experimental_settings}
\end{table*}

\textbf{Korteweg-de Vries Equation.}
The Korteweg-de Vries system, which elucidates the evolution of waves in nonlinear wave phenomena, can be described by the equation:
\begin{equation}
\label{kdveq}
    \frac{\partial u}{\partial t} = -u\frac{\partial u}{\partial x} - \frac{\partial^3 u}{\partial x^3}
\end{equation}
We got the 8 sets of data: 5 for training, 1 for validation, and 2 for testing. The data sets had spatial domain size $x \in [0,64]$, where $\Delta t$ is 15 times $\delta t$ and $\delta t = 0.01s$.

\textbf{2D Burgers Equation.} The 2D Burgers' equation is commonly employed as a benchmark model for comparing and evaluating different computational algorithms, and describes the complex interaction between nonlinear convection and diffusion processes in the way like:
\begin{align}  
\frac{\partial u}{\partial t} &= -u u_x - v u_y + \nu   \nabla^2 u , \\  
\frac{\partial v}{\partial t} &= -u v_x - v v_y + \nu \nabla^2 v  .  
\end{align}  
The $u$ and $v$ are the fluid velocities and $\nu$ denotes the viscosity coefficient. In this case, we choose $\nu = 0.005$, where $\delta t = 0.001s$.

\textbf{2D FitzHugh-Nagumo Equation.} The FitzHugh-Nagumo system can be described by the equation:
 \begin{align}
      \frac{\partial u}{\partial t} &= \mu_u \nabla^2 u+u-u^3-v+\alpha,\\
     \frac{\partial v}{\partial t}&=  \mu_v \nabla^2 v+(u-v)\beta.
 \end{align}
The coefficients $\alpha = 0.01$ and $\beta = 0.25$, governing the reaction process, take distinct values, while the diffusion coefficients are $\mu_u = 1$ and $\mu_v = 100$. In terms of time,  $\delta t = 0.002s$.

\textbf{Gray-Scott Equation.} The Gray-Scott equations describe the temporal and spatial variations of chemical concentrations in reaction-diffusion systems, which can be described by the equation:
 \begin{align}
\frac{\partial u}{\partial t} &= D_u \nabla^2 u - uv^2 + F(1-u) ,\\
\frac{\partial v}{\partial t} &= D_v \nabla^2 v + uv^2 - (F+k)v .
 \end{align}
Here, in the two-dimensional case, $D_u$ and $D_v$ represent the diffusion coefficients of the two substances, with specific values of $D_u = 2.0 \times 10^{-5}$ and $D_v = 5.0 \times 10^{-6}$. $F = 0.04$ denotes the growth rate of the substance, while $k= 0.06$ signifies its decay rate. In the 2D Gray-Scott case, we got 5 trajectories for training, 1 trajectory for validation and 3 trajectories for testing, where $\delta t = 0.5s$. 

In the three-dimensional case, we have the parameters: $D_u=0.2$, $D_v=0.1$, $F=0.025$, and $k=0.055$. We got 3 trajectories for training, 1 trajectory for validation and 2 trajectories for testing, where $\Delta t = 15\delta t$ and $\delta t = 0.25s$.

\section{G: Supplement results}

\begin{table}[htbp]
    \centering

    \label{tab:experiments}
    \begin{tabular}{lcc}
        \toprule
        Models     & RMSE  & MAE   \\
        \midrule
        PIMRL      & \textbf{0.0050} & \textbf{0.0039} \\
        PeRCNN     & 0.0137          & 0.0109          \\
        FNO        & 0.0626          & 0.0502          \\
        FNO-coarse & 0.0415          & 0.0332          \\
        \bottomrule
    \end{tabular}
        \caption{The experiments between PIMRL and baseline models on the Burgers example of FNO benchmark.}
\end{table}

Considering the constraints on the length of the main text, we have placed some important supplementary results in the appendix of the article. First, we conducted additional experiments on the FN equation, expanding our tests to include more advanced baselines such as F-FNO, U-NO, and MWT (shown in table \ref{tab:fn_results}). The experimental results indicate that data-driven methods do not perform well on small training datasets, which further corroborates the effectiveness of the PIMRL proposed in our paper.

 \begin{figure}[!t]
     \centering
    \includegraphics[width=0.4\textwidth]{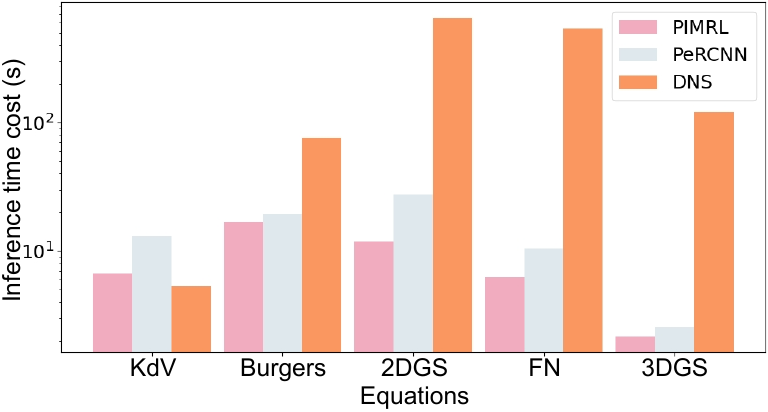}
    \caption{Computational time for comparison}
    \label{fig:filter}
 \end{figure}

\subsection{Inference Time}

As shown in Figure \ref{fig:filter}, PIMRL significantly reduces the computational time cost. Compared to traditional methods such as Direct Numerical Simulation (DNS), our framework is substantially faster, demonstrating a significant improvement in computational efficiency. Additionally, PIMRL not only delivers superior predictive accuracy, but also outperforms PeRCNN in terms of computational efficiency. The numerical methods used for DNS are consistent with the parameter settings during data generation. The difference lies in the required prediction time length, which needs to be the same as that of PIMRL and PeRCNN for a fair comparison. Overall, PIMRL achieves fast and accurate long-term predictions, making it highly advantageous for both precision and real-time applications.

To demonstrate the stability of PIMRL, we have included corresponding error bars in the appendix. These results provide sufficient evidence that our model maintains good stability even when trained with a very limited amount of data. The inclusion of error bars offers a quantitative assessment of the variability in our model's predictions, thereby reinforcing the robustness of PIMRL under data-scarce conditions. 

Furthermore, to further validate the capabilities of our model, we have included an additional dataset, introducing the classic 1D Burgers equation dataset from FNO (Fourier Neural Operator). The experimental results show that our model continues to exhibit significant improvements over other baseline models on this dataset. This not only demonstrates the effectiveness of PIMRL in tackling challenging partial differential equation problems but also highlights its broad applicability and superior performance.

Finally, we provide a comparison of the running time, parameter size, and GPU memory usage between PIMRL and the baseline models. This analysis offers insights into the computational efficiency and resource requirements of our model relative to existing approaches, further highlighting the practical advantages of PIMRL in terms of performance optimization and scalability.

\begin{table}[ht]
\centering

\begin{tabular}{lcc}
\toprule
\textbf{Model} & \textbf{PIMRL} & \textbf{PeRCNN} \\
\midrule
Kdv & $0.0457 \pm 0.0053$ & $0.0942 \pm 0.0082$ \\
Burgers & $0.0068 \pm 0.0006$ & $0.0075 \pm 0.0008$ \\
2DGS & $0.0133 \pm 2.4 \times 10^{-12}$ & $0.0455 \pm 1.9 \times 10^{-11}$ \\
FN & $0.1349 \pm 0.0040$ & $0.1591 \pm 0.0061$ \\
3DGS & $0.0381 \pm 0.0015$ & $0.0532 \pm 0.0027$ \\
\bottomrule
\end{tabular}
\caption{ Results with Error Bar under RMSE metric.}
\label{tab:comparison}
\end{table}
 
\section{H: Limitations and Future Work}

 While our proposed PIMRL framework demonstrates significant promise, we acknowledge several limitations that pave the way for future research. Firstly, the performance sensitivity of the framework to the duration of the data gap between burst-sampling intervals has not been exhaustively studied. Understanding the critical point at which the macro-module's predictive error becomes too large for the micro-module to correct is essential for practical applications. Secondly, our current experiments were conducted on clean, numerically generated data. The framework's robustness to varying levels of noise, which is ubiquitous in real-world sensor measurements, remains an open question for investigation.

Looking ahead, our efforts will address these limitations and extend the framework's capabilities. We plan to explore more advanced architectures for the macro-scale module to enhance both computational efficiency and predictive accuracy, particularly for highly complex dynamical systems. Furthermore, we will work to integrate super-resolution techniques, enabling the model to generate high-fidelity predictions on finer grids than those it was trained on. Systematically evaluating the model's performance under noisy and more sparsely sampled conditions will also be a priority. These advancements will be crucial for unlocking PIMRL's full potential in modeling intricate physical phenomena and broadening its impact across scientific computing.

\end{document}